\documentclass[sigconf]{acmart}
\usepackage{booktabs} % For formal tables
\usepackage{mathtools}
\usepackage{placeins}

\setcopyright{rightsretained}
\acmDOI{10.475/123_4}

\acmConference[WOODSTOCK'97]{ACM Woodstock conference}{July 1997}{El
  Paso, Texas USA} 
\acmYear{2018}
\copyrightyear{2018}

\begin{document}
\title{HeteroMed: Heterogeneous Information Network for Medical Diagnosis}
%\titlenote{Produces the permission block, and
% \subtitle{Extended Abstract}
% \subtitlenote{The full version of the author's guide is available as
%   \texttt{acmart.pdf} document}

\author{Anahita Hosseini}
% \authornote{Dr.~Trovato insisted his name be first.}
\affiliation{%
  \institution{University of California Los Angeles}
  \streetaddress{404 Westwood Plaza}
  \city{Los Angeles} 
  \state{California} 
  \postcode{90095}
}
\email{anahosseini@ucla.edu}

\author{Ting Chen}
% \authornote{The secretary disavows any knowledge of this author's actions.}
\affiliation{%
  \institution{University of California Los Angeles}
  \streetaddress{404 Westwood Plaza}
  \city{Los Angeles} 
  \state{California} 
  \postcode{90095}
}
\email{tingchen@cs.ucla.edu}

\author{Wenjun Wu}
% \authornote{This author is the
%   one who did all the really hard work.}
\affiliation{%
  \institution{University of California Los Angeles}
  \streetaddress{404 Westwood Plaza}
  \city{Los Angeles} 
  \state{California} 
  \postcode{90095}
}
\email{wenjunwu@ucla.edu}

\author{Yizhou Sun}
\affiliation{%
  \institution{University of California Los Angeles}
  \streetaddress{404 Westwood Plaza}
  \city{Los Angeles} 
  \state{California} 
  \postcode{90095}
}
\email{yzsun@cs.ucla.edu}

\author{Majid Sarrafzadeh} 
\affiliation{%
  \institution{University of California Los Angeles}
  \streetaddress{404 Westwood Plaza}
  \city{Los Angeles} 
  \state{California} 
  \postcode{90095}
}
\email{majid@cs.ucla.edu}

\renewcommand{\shortauthors}{A. Hosseini et al.}

\begin{abstract}
With the recent availability of Electronic Health Records (EHR) and great opportunities they offer for advancing medical informatics, there has been growing interest in mining EHR for improving quality of care. Disease diagnosis due to its sensitive nature, huge costs of error, and complexity has become an increasingly important focus of research in past years. Existing studies model EHR by capturing co-occurrence of clinical events to learn their latent embeddings. However, relations among clinical events carry various semantics and contribute differently to disease diagnosis which gives precedence to a more advanced modeling of heterogeneous data types and relations in EHR data than existing solutions.

To address these issues, we represent how high-dimensional EHR data and its rich relationships can be suitably translated into HeteroMed, a heterogeneous information network for robust medical diagnosis. Our modeling approach allows for straightforward handling of missing values and heterogeneity of data. HeteroMed exploits metapaths to capture higher level and semantically important relations contributing to disease diagnosis. Furthermore, it employs a joint embedding framework to tailor clinical event representations to the disease diagnosis goal. 
To the best of our knowledge, this is the first study to use Heterogeneous Information Network for modeling clinical data and disease diagnosis. Experimental results of our study show superior performance of HeteroMed compared to prior methods in prediction of exact diagnosis codes and general disease cohorts. Moreover, HeteroMed outperforms baseline models in capturing similarities of clinical events which are examined qualitatively through case studies.
% \footnote{This is an abstract footnote}
\end{abstract}

\begin{CCSXML}
<ccs2012>
<concept>
<concept_id>10010405.10010444.10010447</concept_id>
<concept_desc>Applied computing~Health care information systems</concept_desc>
<concept_significance>500</concept_significance>
</concept>
<concept>
<concept_id>10010405.10010444.10010449</concept_id>
<concept_desc>Applied computing~Health informatics</concept_desc>
<concept_significance>300</concept_significance>
</concept>
<concept>
<concept_id>10002951.10003317.10003338.10003343</concept_id>
<concept_desc>Information systems~Learning to rank</concept_desc>
<concept_significance>300</concept_significance>
</concept>
<concept>
<concept_id>10002951.10003317.10003338.10003346</concept_id>
<concept_desc>Information systems~Top-k retrieval in databases</concept_desc>
<concept_significance>300</concept_significance>
</concept>
</ccs2012>
\end{CCSXML}

\ccsdesc[500]{Applied computing~Health care information systems}
\ccsdesc[300]{Applied computing~Health informatics}
\ccsdesc[300]{Information systems~Learning to rank}
\ccsdesc[300]{Information systems~Top-k retrieval in databases}

\keywords{Heterogeneous Information Network; Electronic Health Record; Health informatics; Network Embedding}

\maketitle

\section{Introduction}
Electronic Health Records (EHR) provide detailed documented information on various clinical events that occur during a patient's stay in the hospital. Laboratory tests, medications, nurse notes, and diagnoses are examples of heterogeneous types of clinical records. Availability of EHR data in recent years has opened great opportunities for researchers to further explore computer-aided advancements in the healthcare domain. One goal of many existing studies is improving clinical decision making and disease diagnosis.

The diagnostic process involves careful consideration of clinical observations (such as symptoms and diagnostic tests), extraction of relevant information, and more importantly paying attention to their relations. A clinical observation is generally non-specific to a single disease. It is its relation or co-occurrence with other observations that can be indicative of a disease \cite{diagErr}. Moreover, the presence of multiple diseases can cause complexity in observations and their relations. These complexities along with a large amount of information to be analyzed by clinicians make their decisions prone to cognitive error and in many cases suboptimal, which can be very costly and in some cases fatal. A study on 100 diagnostic errors showed that cognitive factor contributed to 74\% of the errors made \cite{diagErr}. Therefore, building a computer-aided diagnosis system is of great importance in reducing error and improving healthcare.

To design such system, obtaining a structured and informative model of the EHR data is necessary. Prior studies employ different approaches for this aim. A group of them employed feature engineering to represent clinical events in EHR and used deep or shallow models for the prediction goals. However, high dimensionality of EHR data, commonness of missing values, and need for extensive clinical knowledge are main challenges that arise in this approach and introduce many limitations. Others employed unsupervised representation learning of clinical events, patients, and visits \cite{unsup:med2vec,unsup:skipgram,unsup:skipgram2,unsup:skipgram3}. These methods that are mostly inspired by Med2vec \cite{unsup:med2vec}, consider co-occurrence of clinical events in different patient records to extract latent embeddings of these entities. However, representations learned are general and not tailored to the goal of diagnosis prediction. More importantly, none of the above-mentioned approaches can capture the rich structure of EHR data and semantics of various relations it contains. This is while some relations make a great contribution to the prediction goal and the model should be able to capture and reflect those into the learned representations. Therefore, any adopted EHR modeling approach should achieve two main goals:
\begin{itemize}
\item Properly capture structure of EHR and semantic of relations
\item Learn representations suitable for disease diagnosis goal 
\end{itemize}
%in medical domain advanced methods not used
To address these requirements and shortcomings of prior models, we propose a  disease diagnosis model based on Heterogeneous Information Network (HIN) \cite{hin1}. HINs, information networks with various types of nodes and relations, have gained lots of attention in recent years due to their ability in distinguishing and learning the different semantics of relations among entities \cite{hinsur} and can be profoundly beneficial to better express the rich network of patients and clinical events in the EHR data.

We introduce how EHR can be translated into an HIN and introduce our node extraction strategies from different formats of data (e.g., raw text, numerical, categorical) present in EHR. We then exploit metapaths \cite{hin1} to introduce composite relation semantics into our network and capture those that are informative for our diagnostic purposes.
We afterward employ a heterogeneous embedding model \cite{heter:mp2vec} to learn representations of clinical events of the network, which samples similar nodes  from pre-defined metapaths. This allows our model to learn similarity of clinical events and patients with respect to semantically important paths rather than random sampling strategy used in prior skip-gram based diagnosis studies. To further tailor latent embeddings to diagnosis prediction goal, a supervised embedding model is jointly learned to adjust representations, using the framework proposed by \cite{main}. While our diagnosis prediction model only utilizes diagnostic information for reasoning the disease, we propose exploiting the treatment information at the time of unsupervised representation learning to improve learned embeddings and capture similarity of clinical events in terms of outcome. Apart from relation-aware modeling and tailored representation learning, HINs also offer the advantage of straightforward handling of missing values, which is a common challenge in clinical data modeling.

We demonstrate that employing HIN for modeling EHR and diagnosis prediction outperforms state of the art models in two levels of general disease cohort and specific diagnosis prediction. We also conduct two case studies to qualitatively reveal the strength of HeteroMed in capturing relations in clinical data which are validated by a clinician. In short, contributions of this study are:
\begin{itemize}
	\item We propose HeteroMed, an HIN-based medical model for disease diagnosis which captures the semantics of clinical relations and learns tailored embeddings for disease diagnosis.
    \item  We demonstrate how EHR data can be translated into an HIN to achieve optimal learning power.
    \item  We empirically show HeteroMed outperforms existing diagnostic models qualitatively and quantitatively.
\end{itemize}
\section{Related Work}
%other prediction task in healthcare —> if time and space
To tackle the problem of disease prediction, initial studies until recent years employed conventional feature engineering methods to extract clinical representations and predict diseases \cite{feature:neural_compare,feature:heartOverview,feature:liver}. However,
feature engineering for clinical domain is a tedious task and requires expert knowledge. Moreover, missing values in EHR pose a great challenge to feature extraction \cite{gen:missingVal}. A number of recent studies \cite{feature:deep,feature:benchmark} employ feature engineering approach along with a deep model to predict high level disease categories, which achieve improvements in results but still experience same issues.

Recognizing discussed challenges, recent studies employ unsupervised representation learning approaches \cite{unsup:pati_sim, unsup:skipgram,future:temporal}. Most of proposed models, inspired by success of word2vec \cite{word2vec1,word2vec2} in natural language processing, turn clinical event records into an ordered sequence of words and employ skip-gram \cite {word2vec2} to capture co-occurrence of clinical events and learn latent embeddings \cite{unsup:skipgram2, unsup:skipgram3}.
To expand the idea, Med2vec \cite{unsup:med2vec} proposes a multilayer representation learning model for clinical code and visit embedding which also initializes the embeddings using skip-gram but modifies them through the network training process. Although these models are successful in eliminating the need for clinical expert knowledge, they fail to capture EHR structure and its internal relations and can only learn a general-purpose representation. It is worth mentioning that the recent studies Dipole \cite{dipole}, RETAIN \cite{retain}, and GRAM \cite{Gram} although similar in subject, are different than ours. These studies are mainly focused on employing history of admissions and diagnoses for future disease prediction. While our study is focused on the diagnosis based on clinical events happening during a single admission.

Heterogeneous Information Networks \cite{hin1} are different from homogeneous ones in their ability of representing multiple types of nodes and relations. This capability has attracted lots of attention in different applications such as personalized recommendation \cite{heter:recom} and malware detection \cite{heter:malware}. Due to the large size of real-world networks and sparsity of data, representing nodes as a low-dimensional vectors is a widely adopted approach in network mining. Network representation learning techniques in general are inspired by word2vec \cite{word2vec1}, among which DeepWalk \cite{deepwalk}, LINE \cite{line}, and Node2vec \cite{node2vec} have been utilized in many network mining researches. Recent studies have adopted similar techniques for heterogeneous network representation learning \cite{heter:emb1,heter:mp2vec, main}. They include the heterogeneity of nodes in the definition of relations and neighbors. Furthermore, it is demonstrated in \cite{main} that a joint embedding approach in heterogeneous node representation learning can lead to improved supervised task performance. In this study, we employ heterogeneous network embedding alongside with the joint learning framework to learn clinical event representations.
\section {Methodology}
In this section, we first put forward the problem definition and terminology used in the study. Then, we introduce how EHR can be viewed as a heterogeneous network and discuss network construction techniques. Lastly, we discuss the training and prediction models adopted for our disease diagnosis task.
\subsection{Problem Definition and Clinical Terminology}
Each record in EHR data is conventionally called a clinical event. A clinical event $e$ can be viewed as a triple: $e= (t, n, v)$ where $t$, $n$, and $v$ respectively denote \textit{type}, \textit{name}, and \textit{value} of it.  \textit{Glucose level of 60} is an example of a clinical event that has type = laboratory test, name = Glucose, and value = 60. Clinical events based on their type may or may not have a value.

Furthermore, the set of clinical event types in EHR is denoted as\\ 
$t_1, t_2,  \dots, t_{|T|} \in T$ where $T = DIAGNOSTIC$  $\cup$  $TREATMENT$ and\\
$DIAGNOSTIC$ = \{\textit{laboratory test}, \textit{symptom}, \textit{age}, \textit{gender}, \textit{ethnicity}, \textit{microbiology test}\} and $TREATMENT$ =  \{\textit{prescription}, \textit{procedure}, \textit{diagnosis}\}. Diagnostic clinical events are the source of information for disease diagnosis. This is while clinical events in treatment category happen after the diagnostic process and should not be directly used for diagnosis prediction.

Therefore, having clinical events for a patient $p$ represented as $E(p) = \{E_{1}(p), \dots, E_{T}(p) \}$ where $E_{t}(p)$ denotes all type $t$ clinical events recorded for $p$, we define the problem of disease diagnosis as prediction of $p$'s diagnosis clinical events ($D_{p}$), denoted as $D_{p} = [E_{t}(p)$ | $t = diagnosis]$, given the diagnostic clinical events of $p$: $\{E_{t}(p)$ | $t \in DIAGNOSTIC\}$. Due to the large size of all possible diagnoses in EHR data, we define  diagnosis prediction as a ranking problem such that top results of prediction model  should ideally match the real diagnosis set ($D_{p}$) for patient $p$.
\subsection{EHR from a Heterogeneous Network Point-of-View}
\begin{figure}
\centering
\includegraphics[width=0.5\textwidth]{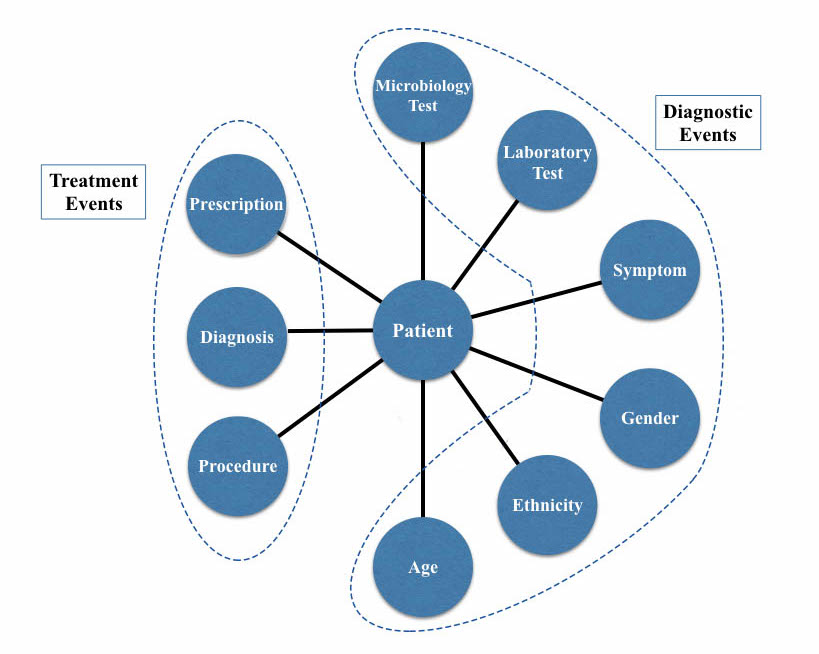}
\caption{EHR heterogeneous network schema.}
\label{fig:network}
\end{figure}

Multiple types of clinical events and their various types of relations can be intuitively viewed as a heterogeneous network.
\begin{definition}
Heterogeneous Information Network is defined as a graph $G = (V, E)$ in which nodes and links between them can have various types. Nodes are mapped to their type by a node mapping function $g_{v} : V \rightarrow A$ where $A$ is the set of all node types and similarly a link mapping function $g_{e} : E \rightarrow R$ maps links to their type where $R$ is the set of all possible link types. By definition we have $|R| > 1$ or $|A| > 1$. Furthermore, $S_{G} = (A, R)$ denotes the network schema.
\end{definition}
%For instance, aspirin and ibuprofen can be two nodes of the type medication.
Different patients and clinical events form the nodes of our clinical heterogeneous network. The type of a node in this network is defined by the type of the clinical event mapped to it. Moreover, links of the network are designed based on the basic EHR relations which are mainly between a patient and a clinical event (e.g., patient's relation to his laboratory tests or symptoms). Figure \ref{fig:network} shows the abstract schema of the network illustrating node types and basic links. The figure also specifies if nodes belong to the treatment or diagnostic type category.

To further enrich the network with semantics of relation in EHR, new compositional relations can be defined using Metapaths \cite{heter:emb1}. 
\begin{definition}
%improve the sentense
Metapaths in HIN define higher order relations between two node types. Having the network schema $S_{G} = (A, R)$, a metapath schema is denoted as $A_{1} \xrightarrow{R_{1}} A_{2} \xrightarrow{R_{2}} ... \xrightarrow{R_{m}} A_{m+1} $.
\end{definition} A metapath is considered as a new link in the network and is added by creating a new connectivity between start and end nodes of any path matching the metapath schema. Metapaths allow our network to better learn the semantics of similarity among nodes. For instance, \textit{patient}$\rightarrow$ \textit{symptom} $\leftarrow \textit{patient}$ captures similarity of patients in terms of their symptoms. 
\subsection{Construction of HIN from Clinical Events}
\label{sec:cons}
%fix this
In this section, we introduce the proper modeling approach for construction of HIN from EHR data and our technique in extraction of some clinical events from raw text.

In general, having a clinical event $e = (t,n,v)$, it can be mapped into a node of type $t$ with identification of $(n,v)$ (e.g., node (Glucose, 60) with laboratory test type). However, in many cases, different values of clinical events with the same type and name convey identical semantic in terms of disease prediction. For instance, various numerical measurements in a laboratory test are considered the same as long as they fall into one of the normal or abnormal ranges. Therefore, a proper modeling strategy should map clinical events with duplicate diagnostic semantic into the same node as failing to do so can negatively affect the power of the model in capturing similarity of nodes. Having this in mind, following steps are taken for mapping clinical events to nodes. Procedure and diagnosis clinical events are mapped based on the icd-9 coding system \cite{icd9}. For each laboratory test, its name coupled with a reported flag which can be either normal or abnormal is considered as a unique node. The same strategy is employed when dealing with microbiology tests, where flags can be sensitive, resistant, or intermediate. Moreover, the age of patients is classified with threshold 15, 30, and 64  based on a statistical analysis of adverse events in different age groups studied by \cite{ageGroup}. Finally, gender, ethnicity, and prescription, which are categorical events, are easily mapped by their unique category names. 
\subsubsection*{\textbf{Symptom Extraction}}
For extraction of symptoms that are commonly found inside raw-text clinical notes, we employed Autophrase \cite{autophrase} which is a novel phrase mining technique that learns high-quality phrases from a large corpus and allows for incorporating domain-specific knowledge bases for achieving highly domain-relevant results. We feed Autophrase with a pool of clinical phrases that are generated from two main sources. Firstly, Medical Subject Headings (MeSH) \footnote{https://www.nlm.nih.gov/mesh/} vocabulary treasure which contains 90,000 medical entry terms and secondly the ICD-10 \footnote{http://www.who.int/classifications/icd/en/} medical coding database. Quality phrases for symptoms are extracted from MeSH "signs and symptoms" category, code C23 and ICD-10 Chapter XVIII. We also run some final filtering steps on results of Autophrase to drop phrases that include measurements, adverbs, or symbols as they do not contribute to our diagnosis goal.

Having all nodes constructed, their connections are added based on simple paths in Figure \ref{fig:network} and selected metapaths discussed in following sections. One of the advantages of HIN is that missing values in EHR only lead to the absence of some links and does not require further management.
\subsection{Heterogeneous Network Embedding for Clinical Events}
\label{sec:unsup}
Given the rich clinical information network, learning a latent and low-dimensional embedding of clinical events that can capture their internal relations is greatly beneficial for further analysis tasks. Inspired by the success of skip-gram \cite{word2vec2} in learning latent word embeddings from the context of words in a corpus, most of homogeneous network embedding techniques \cite{line,node2vec} rely on neighbor prediction paradigm. In this approach, given a network $G = (V, E)$, and an embedding function $f: V \rightarrow R ^ {d}$ that maps each node to a $d$ dimensional vector, the objective is to maximize the probability of observing neighborhood of a node $v$, denoted as $N(v)$, conditioned on $v$ \cite{node2vec}.
\begin{displaymath}
\underset{f}{\operatorname{argmax}} 
\prod_{v \in V}\prod_{c \in N(v)} Pr(c | v)
\end{displaymath}
where the probability $Pr(c | v)$ is defined as a softmax function normalized with respect to all network nodes.

To exploit the rich structural information of EHR data and enrich semantics of similarity among different nodes, we employ an extension of above paradigm to heterogeneous networks that incorporates variety in node types and metapaths in the definition of node neighborhood and the objective function \cite{main,heter:mp2vec}. In particular, with presence of multiple node types, neighborhood of a node $v$ is defined as $N (v) = \{N_{1}(v), N_{2}(v), ..., N_{T}(v)\}$ where $N_{t}(v)$ denotes type $t$ neighbors of $v$ and $T$ is the number of node types.

Moreover, having multiple types of path leaving a node (simple or metapath), the neighbor prediction probability function $Pr(c|v)$ should be also conditioned on the type of path used. Specifically, the probability of visiting a neighbor $c$ of a node $v$ under path r with schema $V_1\rightarrow ... \rightarrow V_l$, is defined as:
\begin{displaymath}
Pr(c| v, r) = \frac{exp(f(c).f(v))}{\sum_{u \in V_{l}} exp(f(u).f(v))}
\end{displaymath}
As computation of above probability is very expensive in large networks, negative sampling \cite{word2vec2} is employed to achieve following objective function:
\begin{displaymath}
Pr(c| v, r) = \log \sigma(f(c).f(v)) + \sum_{1}^{m} \mathbb{E}_{u_{l}\sim P_{l}(u_{l})} \log \sigma(-f(u_{l}).f(v))
\end{displaymath}
where $m$ negative sample nodes are drawn based on their node degree and from nodes having the same type as $r$ destination type ($V_{l}$). Therefore, a training step randomly samples a path schema $r$ and two nodes $v$ and $c$ connected under $r$, along with $m$ negative sampled nodes and employs SGD to update their embeddings. The tuple ($v$, $c$) is sampled based on the normalized number of links under the path $r$ over each node tuples.

Although treatment clinical events should not be directly used in the diagnosis prediction, they can be profoundly beneficial in the unsupervised embedding model for capturing similarity of diagnostic clinical events in terms of consequent treatment. For instance, by including prescription and the metapath $symptom \leftarrow patient \rightarrow prescription$, into the embedding model, it can learn similarity among symptoms that lead to the same prescription. Therefore, for training the unsupervised embedding model, we first select the set of advantageous treatment nodes to be added to the embedding model by evaluating the performance-gain obtained from each or combination of them. Next, among many possible metapaths, we select candidate paths mainly from those that link a diagnostic event to a treatment one through a patient (such as the one above). We also compare candidate metapaths in terms of performance-gain when they are added to the network separately and incrementally and select the best configuration.
\subsection{Diagnosis Prediction}
\label{sec:sup} 
\begin{figure}
\centering
\includegraphics[width=0.5\textwidth]{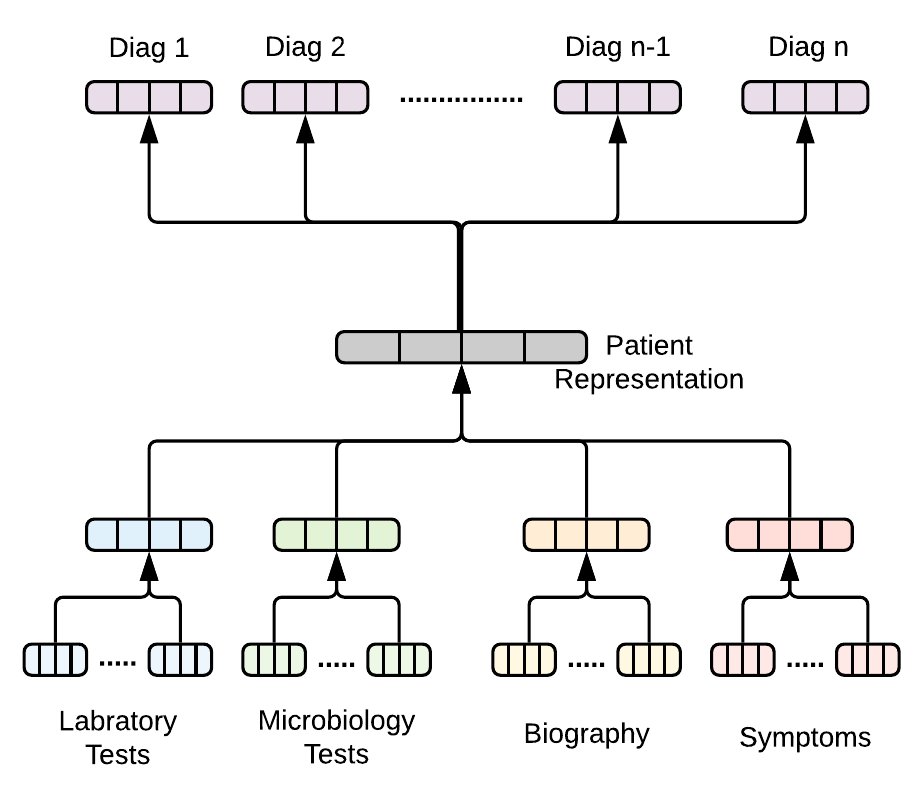}
\caption{Diagnosis prediction flow.}
\label{fig:sup}
\end{figure}

When node embeddings are present, the process of diagnosis prediction for a new patient involves construction of the patient's representation based on his clinical events and ranking diagnosis codes according to their dot product similarity to the patient's representation. Figure \ref{fig:sup} shows the overview of the prediction flow. Given a patient $p$, his type $t$ neighborhood ($N_{t}(p)$) can be summarized into a latent embedding ($f_{t}(p)$) by averaging its members:
\begin{displaymath}
f_{t}(p) = \sum_{n \in N_{t}(p)} \frac{f(n)}{|N_{t}(p)|}
\end{displaymath}
Having clinical events of $p$ grouped into latent type embeddings ($f_{t}(p)$), a representation for $p$ can be intuitively achieved by aggregating them, but with different weights for each type ($w_{t}$) to capture importance of the type in diagnosis prediction.
\begin{displaymath}
f(p) = \sum_{t} w_{t}f_{t}(p)
\end{displaymath}
Finally, a diagnosis $d$ is scored and ranked by a dot product similarity between $p$ and $d$ embeddings: $s(d, p) = f(d).f(p)$. 
\subsection{Tailoring Node Representations to Diagnosis Prediction}
The heterogeneous network embedding model discussed in section \ref{sec:unsup}, does not have a direct guidance for learning representations that are specifically suitable for disease diagnosis aim and learns a general knowledge of the network. To add such guidance and provide diagnostic knowledge to the model, following \cite{main} we employ the diagnosis prediction flow discussed in section \ref{sec:sup} as a supervised embedding process and jointly use with the unsupervised model at the time of representation learning to tailor embeddings to disease diagnosis goal.

Recalling computation of prediction score from section \ref{sec:sup}, which is defined for a tuple of diagnosis $d$ and patient $p$, we have:
\begin{displaymath}
s(d, p) = f(d).f(p) = f(d)\sum_{t} w_{t} f_{t}(p) =
\end{displaymath}
\begin{displaymath}
f(d)\sum_{t} w_{t}\left( \sum_{n \in N_{t}(p)} \frac{f(n)}{|N_{t}(p)|} \right)
\end{displaymath}
We can employ a hinge loss ranking objective for the triple ($p$, $d$, $\sim d$) to update node embeddings ($f$) and node type weights ($w_{t}$).
\begin{displaymath}
max (0, -s(d,p) + s(\sim d,p) + \sigma)
\end{displaymath}
where $d$ and $\sim d$ are positive and negative sampled diagnosis for $p$ and scores $s(d, p)$ and $s(\sim d,p)$ are calculated for them respectively. 
To jointly learn embeddings, objectives of the two supervised and unsupervised models are combined to form the joint objective as:\\
\begin{displaymath}
\mathbb{Z}_{joint}= \omega.\mathbb{Z}_{unsupervised} + (1 - \omega). \mathbb{Z}_{supervised} + \lambda \sum_{n} \Vert f(n)\Vert_{2}^{2}
\end{displaymath}
where $\omega \in [0,1]$ is a pre-defined parameter for tuning importance of either models and a regularization term is added to prevent over-fitting of learned representations.

Therefore, a training step in the joint representation learning model works as follows. We draw one of the embedding models based on $Bernoulli(\omega)$. If the unsupervised model is drawn, its objective function is used on a mini-batch of randomly drawn triples ($r$, $v$, $c$) and $m$ negative samples to update representations. Otherwise, the supervised objective is used for a mini-batch of drawn triples ($p$, $d$, $\sim d$) to update type weights $(w_t)$ and representations ($f$). Negative samples are drawn in both cases from a unigram distribution based on node degree \cite{word2vec2}. 
\section{Experiments}
In this section we evaluate HeteroMed through three sets of experiments. First, it is evaluated under different design configurations. Then its diagnosis prediction performance is compared to various baseline models and finally it is quantitatively evaluated through two case studies.
\subsection{Dataset}
Experiments of this study are conducted on the publicly available Medical Information Mart for Intensive Care III (MiMIC III) \cite{mimic} dataset. It contains a comprehensive clinical data for forty thousand patients admitted to the ICU department of BIDMC hospital during 11 years. The MIMIC dataset is organized into 26 tables containing clinical event records for each admission to the ICU and other general information such as definitions of clinical terms.  Table ~\ref{tab:tabels} lists utilized database tables alongside with main columns used and a short description for each table.
\begin{table*}
  \caption{MIMIC tables used in this study.}
  \begin{tabular}{l l l}
    \toprule
    Table name & Main Columns &  Description\\
    \midrule
    patients\_icd & gender, DOB, ethnicity  & Name and demographic information of patients\\ \hline
    procedures\_icd & icd9\_code  & Procedure events such as brain monitoring, tubing, injection\\ \hline
    prescriptions & generic\_drug\_name & Drugs prescribed in each admission\\ \hline
    microbiologyevents & spec\_itemid, interpretation  & Microbiology tests and their sensitivity level; eg. fungi, bacteria \\ \hline
    labevents & itemid, flag  & Laboratory results and their flag (normal, abnormal); eg. Blood Glucose\\ \hline
    Diagnosis\_icd & icd9\_code  & Prescribed diagnosis codes.  \\ \hline
    noteevents & Category = \" Discharge Summary\"  & Raw text notes recorded by nurses which includes symptoms and other \\
    &&clinical information collected on admission time.\\
  \bottomrule
\end{tabular}
  \label{tab:tabels}
\end{table*}
In this study, each admission of an adult subject (aged 15 years or older) to the hospital is considered as a sample and called a \textit{patient stay}. Few subjects with multiple ICU stays in a single hospital admission were excluded due to the insufficiency of diagnosis information provided for them in MIMIC. 

Following these steps, we obtained a sample set of 46,641 patient stays from which 10,000 were randomly sampled for the test set and 36,641 remaining for the training set. The heterogeneous network was then constructed with the strategy discussed in section \ref{sec:cons} using our train set. Table ~\ref{tab:stats} lists statistical details for nodes of this network.
\begin{table}
  \caption{Node statistics for the HIN network.}
  \begin{tabular}{l c c c}
    \toprule
    Node Type & abbreviation &Train & Test\\
    \midrule
    Patient stay & pati & 36641 & 10000\\
    Procedures & proc & 1673 & 746\\
    Prescription & pres  & 6000 & 3523\\
    Microbiology & micro & 212 & 63\\
    Laboratory & lab  & 1870 & 1045\\
    Diagnosis & diag & 5605 & 2745\\
    Symptom & symp & 1602 & 435\\
    Gender & gen & 2 & 2\\
    Age group & age  & 3 & 3\\
    ethnicity & eth  & 40 & 32\\
  \bottomrule
\end{tabular}
  \label{tab:stats}
\end{table}
Furthermore, in addition to 9 length one basic links of the network, 9 other  candidate metapaths were selected. Table ~\ref{tab:stats2} lists both types of paths with their frequency in the constructed network. As patient node is a central hub in our metapaths, each path is denoted only by its start and end node types. (e.g., lab-symp denotes the \textit{laboratory test}$\leftarrow$ \textit{patient} $\rightarrow$ \textit{symptom} metapath).
\begin{table}
  \caption{Edge statistics for constructed network.}
  \begin{tabular}{l c | l c}
    \toprule
    simple links & count & metapaths & count\\
    \midrule
    pati-proc & 94,452 & lab-diag & 1,155,278\\
    pati-pres & 757,195  & symp-diag & 261,861\\
    pati-micro & 19,768 & lab-proc & 341,907\\
    pati-lab & 1,948,360 & lab-pres & 770,297\\
    pati-age & 36,641 & symp-pres & 223,666\\
    pati-diag & 292,473 & symp-proc & 63,424\\
    pati-symp & 307,325 & lab-symp & 214,356\\
    pati-gen & 36,641 & micro-lab & 14,394\\
    pati-eth & 35,342 & micro-symp & 8,696\\
  \bottomrule
\end{tabular}
  \label{tab:stats2}
\end{table}
\subsection{Evaluation Strategies and Implementation Details}
Disease diagnosis is conducted in two levels in this study. First, exact diagnosis code prediction as a ranking problem and second general disease cohort prediction as a multi-label classification problem which are evaluated with \textit{MAP@k} and \textit{AUROC score} respectively. MAP@k is a metric widely employed in information retrieval and reports the mean of average precision at k (AP@k) over all test samples. In this study, having a ranked diagnoses list returned by the prediction model, AP@k shows the averaged precision over all the positions in the list that the diagnosis is correct and has index less than k.

AUROC is a goodness of binary prediction metric based on different cut-off thresholds on classifier prediction score. Here, AUROC is computed for each of disease cohorts based on the scores computed by our supervised prediction model and baselines for each cohort. 

For training HeteroMed and learning node embeddings, a mini-batch of 500 patients has been used at each training step with embedding vector size of 128. The unsupervised embedding model is selected with 4 times higher probability than the supervised model when performing the joint representation learning. Furthermore, each step of unsupervised approach draws 100 negative diagnosis samples for each patient based on the diagnosis node degree. 
\subsection{Evaluation of Proposed Method}
In this section, we demonstrate experimental results of evaluating performance of HeteroMed under different metapaths and node selection configurations. 
\subsubsection*{\textbf{Treatment Node Selection}}
\begin{figure}
\centering
\includegraphics[width=0.5\textwidth]{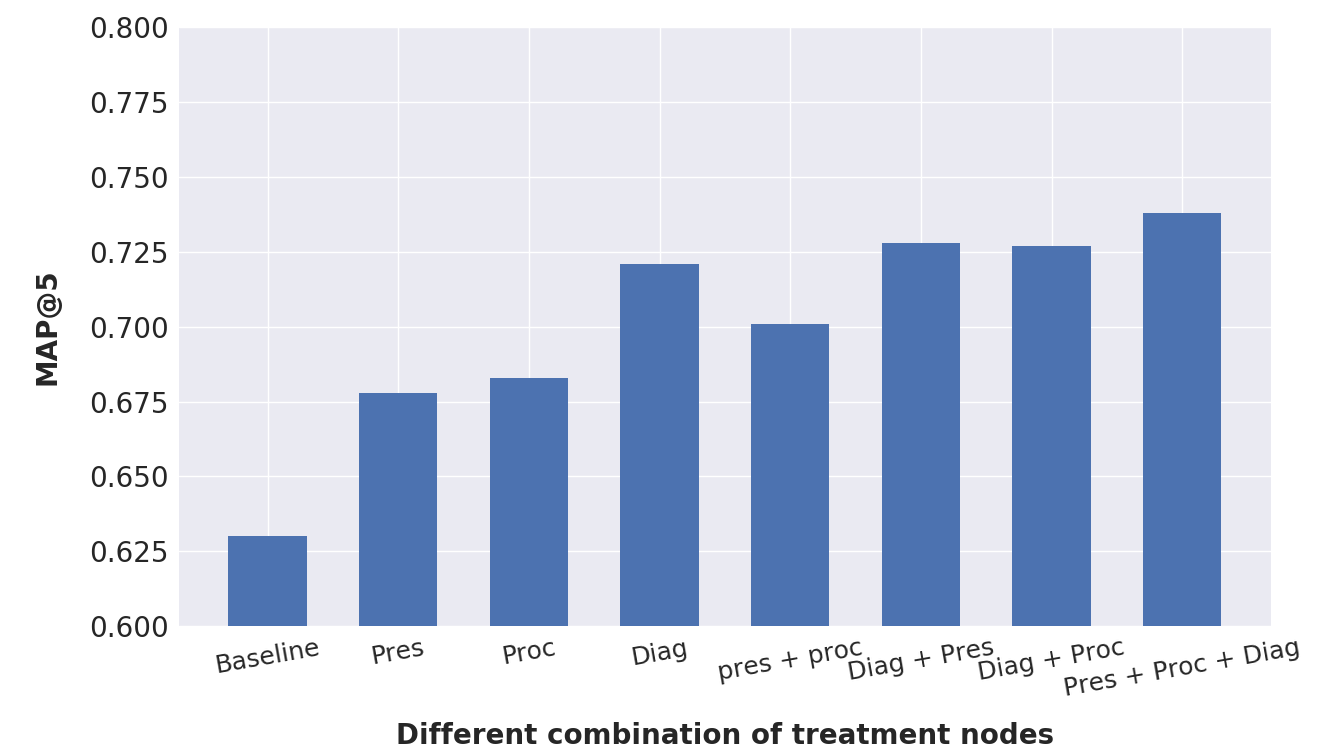}
\caption{Treatment node selection evaluation.}
\label{fig:features_eval}
\end{figure}

\mbox{}\\
In this part, we evaluate performance-gain obtained, when each or a combination of treatment nodes (proc, pres, diag) are included in the network of unsupervised embedding model. Figure ~\ref{fig:features_eval} illustrates the comparison to the baseline performance in which the network only contains diagnostic nodes.

We can observe that among treatment nodes, "diagnosis" show a great advantage to be added to the unsupervised embedding process. This is partly due to the fact that any improvement in diagnosis node embeddings directly impacts performance of diagnosis prediction. Procedure and prescription nodes also impact the performance in a positive way. This is while they are not included in the prediction step of diagnosis. Results of this experiment confirm the advantage of utilizing the whole set of available information in the unsupervised representations learning process.
\subsubsection*{\textbf{Metapath Selection}}
\begin{figure}
\centering
\includegraphics[width=0.5\textwidth]{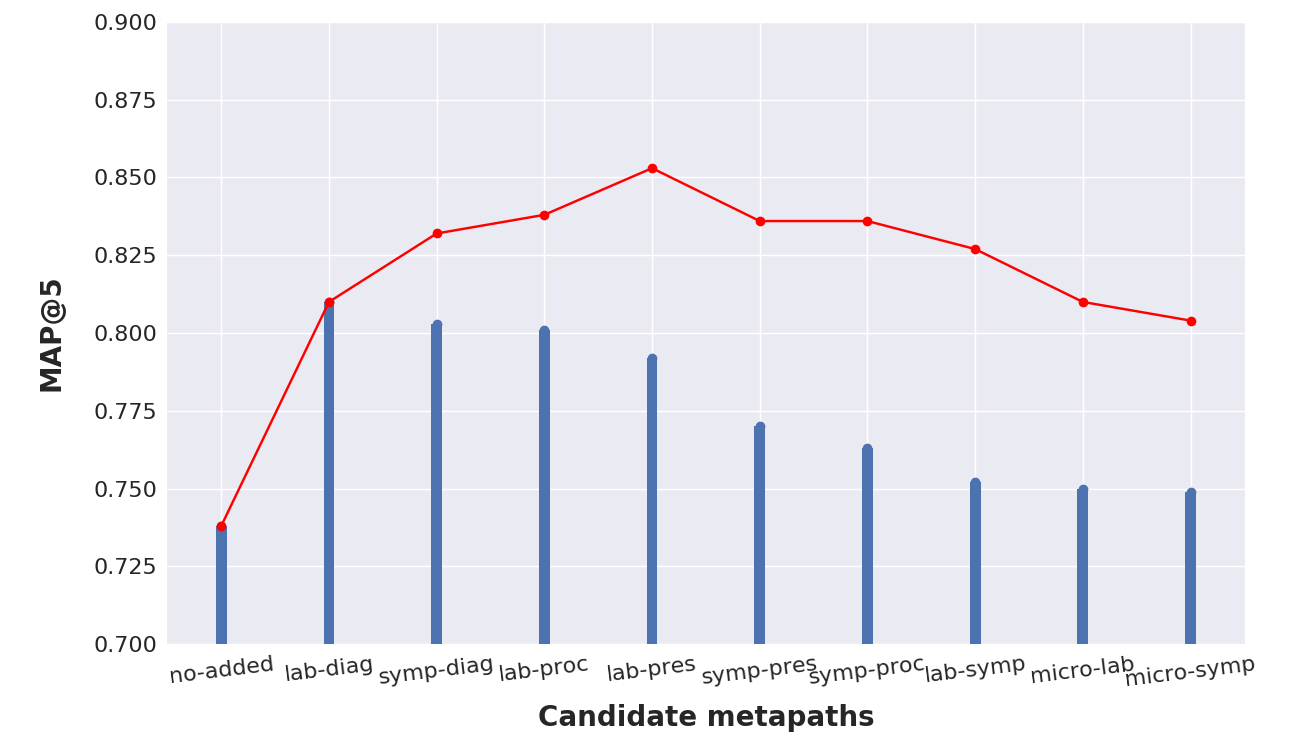}
\caption{Metapath selection evaluation. 
Red line denotes additive performance and blue bars denote single path performance.}
\label{fig:metapath_eval}
\end{figure}

\mbox{}\\
In the second part of this experiment, we evaluate performance-gain obtained by using selected metapaths listed in Table ~\ref{tab:stats2}. The results are elaborated in Figure \ref{fig:metapath_eval}. The blue bars in the figure show the performance for each metapath when added separately to the baseline and are sorted in descending order based on this measure. The red line, however, evaluates performance when these paths are accumulated incrementally to the model. Results of this experiment indicate that the combination of 4 first metapaths (lab-diag, symp-diag, lab-symp, lab-pres) provides us with the optimal performance for the disease diagnosis goal. This is while adding more paths leads to a gradual performance drop. This observation further clarifies the significant advantage of metapath-based neighbor sampling rather than the random neighbor sampling used in prior medical domain studies.

Based on these results, the model used in all succeeding experiments employs all treatment nodes and the 4 above-mentioned metapaths in its representation learning process.

\subsection{HeteroMed Compared to Other Diagnostic Models}
To further assess our model, we compare its diagnosis performance to selected state of the art models in two levels. First, when exact disease codes are to be predicted and second when disease cohorts are desired. 
\subsubsection*{\textbf{Exact Code Prediction}} ~\\
In this experiment, we try to rank exact disease codes for a patient stay. In this part, we compare HeteroMed only to embedding based models as the size of diagnosis codes is too large to be predicted by a supervised classifier. The baseline models in this task are: 
\begin{description}
    \item \textbf{Med2vec}: Med2vec \cite{unsup:med2vec} is a multilayer medical embedding neural network which learns embeddings of medical events and visits using an approach inspired by word2vec. We modified the last softmax layer of this model to predict diagnosis codes for the current visit as the model originally predicted disease codes for last or future visits. This method is implemented by Theano python library \cite{theano} and representation sizes are chosen to be 100.
	\item \textbf{Skipgram-embedded}: We use word2vec (skip-gram) to learn network node representations similar to the way it is employed in prior studies. In particular, all clinical events associated with an admission are considered as words and are concatenated to form sentences. The window size is set to the maximum length of sentences so that all clinical events (words) in an admission (sentence) can be sampled as neighbors. The method is implemented by the open source python tool, Gensim \cite{gensim}. Node representations learned are fed into the supervised prediction model (section \ref{sec:sup}) to score diagnoses and rank them.
    \item \textbf{HeteroMed-embedded}: We learn node embeddings by only employing the unsupervised representation learning approach introduced in section \ref{sec:unsup}. We use the same set of metapaths as the main model for this aim. As the previous method, the supervised diagnosis model is then used to rank diagnoses.
\end{description}
\begin{figure}
\centering
\includegraphics[width=0.5\textwidth]{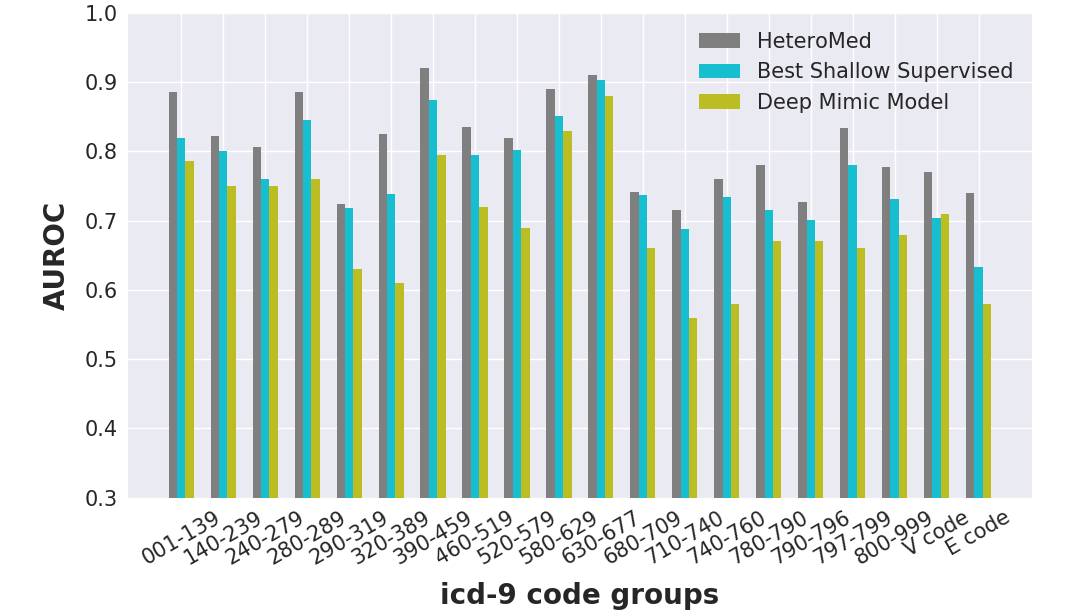}
\caption{Disease cohort prediction evaluation.}
\label{fig:sup_eval}
\end{figure}
\begin{table}
  \caption{Comparison of HeteroMed model to baselines for exact code prediction.}
  \begin{tabular}{l c c c}
    \toprule
    Model Name & MAP@3 & MAP@5 & MAP@10\\
    \midrule
    Med2vec  & 0.75 & 0.78 & 0.79 \\
    Skipgram-embedded & 0.73 &0.76 & 0.77 \\
    HeteroMed-embedded & 0.78 & 0.79 & 0.80 \\
    HeteroMed & \textbf{0.81} & \textbf{0.85} & \textbf{0.87} \\
  \bottomrule
\end{tabular}
  \label{tab:result:comp}
\end{table}
\begin{table*}
  \caption{Similarity search results.}
  \begin{tabular}{c @{\extracolsep{1pt}} c @{\extracolsep{12pt}} c @{\extracolsep{1pt}} c @{\extracolsep{12pt}} c @{\extracolsep{1pt}} c}
    \toprule
     \multicolumn{2}{c}{Diabetes }& \multicolumn{2}{c}{Cold} & \multicolumn{2}{c}{Anemia (lack of blood)}\\ 
     \cmidrule{1-2} \cmidrule{3-4} \cmidrule{5-6}
     HeteroMed  & skipgram                        & HeteroMed   & skipgram                   & HeteroMed   & skipgram \\ \midrule
    %0 & diabetes & diabetes                        & anxiety & anxiety\\
     \textbf{peripheral neuropathy} & \textbf{dietary change}     & \textbf{general pain} & \textbf{fever}        & \textbf{fatigue} & weight loss\\ 
     
    \textbf{sleep apnea}  & tightness              & \textbf{fever} & \textbf{sick contact}                &\textbf{malaise} & \textbf{allergy reaction to iron}\\
    
    \textbf{leg tingling} & confusion                   & \textbf{chill} & constipation                & \textbf{heart palpitation} & penile discharge\\
    
    \textbf{urinary frequency} & speak difficulty       & \textbf{sore throat} & \textbf{muscle pain}          & \textbf{itchy skin} & sick contact\\
    
    \textbf{ulcers} & nausea                            & swelling & \textbf{recent travel}    & \textbf{bloody stool} & \textbf{shortness of breath}\\
    
    \textbf{dietary change} & \textbf{rash}                      & \textbf{allergy} & limb pain                 & bruising & stuffy nose\\
    
    \textbf{burning} & fever                         &\textbf{tightness}  & urinary changes          & \textbf{abdominal pain} & \textbf{leg tingling}\\
    
    abdominal pain & mental status change      & \textbf{sinus congestion} & \textbf{cough}              & \textbf{nausea} & suicidal attempt\\
    
    \textbf{thirst} & \textbf{numbness}             & \textbf{cough} & stiff neck & \textbf{chills} & \textbf{abdominal pain}\\
    
    \textbf{itchy skin} & sleepiness                      & blurred vision &  \textbf{runny nose}     & \textbf{cramps} & \textbf{jaundice}\\
        %pain\\& chills &  abdominal
  \bottomrule
\end{tabular}
  \label{tab:result:case1}
\end{table*}
\subsubsection*{\textbf{Disease Cohort Prediction}} ~\\
icd-9 diagnosis coding provides 20 code groups that correspond to 20 high level disease cohorts. In this part, we aim to predict all disease cohorts that a patient is diagnosed with. Having multiple diagnosis codes for a patient stay, different groups of diseases may be involved which turns the problem into a multi-label classification. When training our model for cohort prediction, only 20 disease nodes are constructed for the network and each disease code of patient is mapped to one of these nodes. Furthermore, in prediction time, scores for all 20 diagnosis nodes are computed to be evaluated. The baseline models are listed below:
\begin{description}
	\item \textbf{Shallow supervised models}: We use feature engineering along with common shallow models, from which Random Forest provided best results. We extracted the same features suggested by \cite{feature:benchmark} but only from tables used to construct our network. We employ Scikit-learn \cite{scikit} for implementation of the basic models.
	\item \textbf{Deep Mimic Model}: We finally compare our results to the ones from mimic learning model \cite{feature:deep} which employs a deep neural network alongside with a Gradient Boosting Model for prediction of icd-9 diagnosis code groups. 
\end{description}
\subsubsection*{\textbf{Results}}\mbox{}\\
The exact code prediction evaluation is depicted in Table ~\ref{tab:result:comp}.
As the results suggest, HeteroMed outperforms all the baseline models in exact diagnosis prediction. The out performance of HeteroMed-embedded model compared to skipgram-embedded model, reveals superiority of relation-aware embedding approach employed in this study to the skip-gram used in conventional clinical models. Furthermore, the Med2vec model outperforms the Skipgram-embedded model although they are both trained based on skip-gram embedding. This can be due to the fact that Med2vec incrementally updates the embeddings with back propagation in its model. However, it sill falls behind HeteroMed that employs relation-aware embedding approach.

Results of the disease cohort prediction are illustrated in Figure ~\ref{fig:sup_eval}. We can observe that HeteroMed performance exceeds baseline models in almost all code groups. In general, performance in some groups are lower than the others which generally corresponds to those diagnosis groups that are sparser in the MIMIC dataset.
\subsection{Case Studies}
%improve this!
In this section, we qualitatively evaluate modeling of EHR data using HeteroMed and validate sensibility of learned clinical event representations. First, we perform a similarity search to find relevant symptoms to three common diseases. We then review results of a sample prediction case. In both experiments, we compare the results to the Skipgram-embedded model introduced in the last section.

Table ~\ref{tab:result:case1} lists top ten related symptoms and observations to three common clinical conditions: Diabetes, Cold, and Anemia. A dot product similarity has been employed to generate these results. To achieve better vision for comparison, results are validated by a clinical expert and relevant symptoms are shown in bold format. Recognizing the fact that symptoms can have hidden and complex relations to diseases, only directly related symptoms to each condition are considered as relevant. 

Results of this experiment confirm the validity of learned representations by our model. Moreover, we can easily observe that HeteroMed ranks relevant symptoms higher than the Skipgram-embedded model and is vividly stronger in understanding relations of symptoms to diseases. One may notice that the intersection among results of two models is small. The large number of symptoms and the fact that a single complication can be described in multiple ways are the main reasons for this observation. For instance, leg tingling, numbness, and peripheral neuropathy can all refer to a similar complication caused by diabetes. 

Table ~\ref{tab:res:case2} shows a sample admission with 9 real diagnosis codes along with the 9 top-ranked predicted codes by each model. Wrong predictions are denoted by a star sign on the top right corner. Furthermore, the main category of each real disease code is specified to provide better understanding of them.
\begin{table}
  \caption{Comparison of sample prediction results for a patient and real diagnosis codes.}
  \begin{tabular}{c c  c c}
    \toprule
    Real Codes & Category & Skipgram & HeteroMed\\
    \midrule
    4282 & Circulatory system & 2875     & 4273  \\
    4254 & Circulatory system & 3970     & 4282  \\
    2875 & Blood organs       &$6841^ *$  & 4583  \\
    4273 & Circulatory system & 281     & $2832^*$ \\
    3970 & Circulatory system & $7217^ *$ & 2875  \\
    5303 & Digestive system   & $427^ *$   & 4254  \\
    4280 & Circulatory system & 4583     & $530^*$  \\
    281  & Blood organs       & 4273     & $260^*$ \\
    4583 & Circulatory system & $2501^ *$ & 281 \\
    
  \bottomrule
\end{tabular}
  \label{tab:res:case2}
\end{table}
The two methods rank a number of wrong codes in their first 9 predictions. However, the superior performance of HeteroMed is noticeable in two aspects. Firstly, we can observe that it is able to detect all disease categories although not predicting exact codes. Specifically, it predicts a more general code of 530 rather than 5303 but both correspond to digestive system diseases. This is while Skipgram-embedded model misses this disease category in its predictions. Secondly, all the codes predicted by HeteroMed belong to the category of diseases that are present in ground truth. However, Skipgram-embedded model predicts codes such as 7217 which belongs to connective tissue diseases.

In general, we can observe that HeteroMed can achieve superior results in major prediction experiments.

\section{Conclusion and Future Work}
In this paper, we study the problem of disease diagnosis from a patient's diagnostic records available in EHR data. We propose modeling of clinical events as a heterogeneous information network, HeteroMed, to address shortcomings of previous methods pursuing same goals. Existing studies ignore the rich structure and relations in EHR data when learning representations of clinical events. HeteroMed is capable of capturing informative relations for the diagnosis goal and use the best relation sampling strategy when learning clinical event representations. It also allows for easy handling of missing values and learning embeddings tailored to the disease prediction goal using a joint embedding framework. Result of our study shows that HeteroMed can achieve significantly better results in diagnosis task and finding clinical similarities. This in turn confirms the benefits of employing heterogeneous information network in modeling clinical data. \\
Future work includes modeling more diverse type of information using heterogeneous network such as timeseries data and joint suggestion of disease and treatment based on available diagnostic information. 
\FloatBarrier

\bibliographystyle{ACM-Reference-Format}
\bibliography{sample-bibliography} 

\end{document}